%% file: conference_101719.tex
\documentclass[conference]{IEEEtran}
\IEEEoverridecommandlockouts
\usepackage{cite}
\usepackage{amsmath,amssymb,amsfonts}
\usepackage{algorithmic}
\usepackage{graphicx}
\usepackage{textcomp}
\usepackage{xcolor}
\def\BibTeX{{\rm B\kern-.05em{\sc i\kern-.025em b}\kern-.08em
    T\kern-.1667em\lower.7ex\hbox{E}\kern-.125emX}}
\usepackage[acronym]{glossaries}

\usepackage{amsmath,stmaryrd,pict2e,picture}

\makeatletter
\newcommand{\pictslash}[2]{%
  \vcenter{\hbox{%
    \sbox0{$\m@th#1\varobslash$}\dimen0=.55\wd0
    \pictslash@aux#2%
  }}%
}
\newcommand{\pictslash@aux}[2]{%
    \begin{picture}(\dimen0,\dimen0)
    \roundcap
    \put(0,#1\dimen0){\line(1,#2){\dimen0}}
    \end{picture}%
}
\makeatother

\input{acronyms}
\begin{document}

\title{UnrealROX+: An Improved Tool for Acquiring Synthetic Data from Virtual 3D Environments\\
%{\footnotesize \textsuperscript{*}Note: Sub-titles are not captured in Xplore and should not be used}
\thanks{}
}

\author{\IEEEauthorblockN{Pablo Martinez-Gonzalez\IEEEauthorrefmark{1},
Sergiu Oprea\IEEEauthorrefmark{1},
John Alejandro Castro-Vargas\IEEEauthorrefmark{1},\\
Alberto Garcia-Garcia\IEEEauthorrefmark{3},
Sergio Orts-Escolano\IEEEauthorrefmark{2},\\
Jose Garcia-Rodriguez\IEEEauthorrefmark{1} and
Markus Vincze\IEEEauthorrefmark{4}}
\IEEEauthorblockA{\IEEEauthorrefmark{1} Department of Computer Technology, University of Alicante, Spain \\
Email: \{pmartinez, soprea, jacastro, jgarcia\}@dtic.ua.es}
\IEEEauthorblockA{\IEEEauthorrefmark{2} Department of Computer Science and Artificial Intelligence, University of Alicante, Spain \\
Email: sorts@ua.es}
\IEEEauthorblockA{\IEEEauthorrefmark{3} Institute of Space Sciences (ICE-CSIC), Spain \\
Email: garciagarcia@ice.csic.es}
\IEEEauthorblockA{\IEEEauthorrefmark{4} Vision for Robotics Laboratory, TU Wien, Austria \\
Email: vincze@acin.tuwien.ac.at}}

%\author{\IEEEauthorblockN{1\textsuperscript{st} Pablo~Martinez-Gonzalez}
%\IEEEauthorblockA{\textit{Department of Computer Technology} \\
%\textit{University of Alicante}\\
%Alicante, Spain \\
%pmartinez@dtic.ua.es}
%\and
%\IEEEauthorblockN{2\textsuperscript{nd} Sergiu~Oprea}
%\IEEEauthorblockA{\textit{Department of Computer Technology} \\
%\textit{University of Alicante}\\
%licante, Spain \\
%oprea@dtic.ua.es}
%\and
%IEEEauthorblockN{3\textsuperscript{rd} John~Alejandro~Castro-Vargas}
%\IEEEauthorblockA{\textit{Department of Computer Technology} \\
%\textit{University of Alicante}\\
%Alicante, Spain \\
%jacastro@dtic.ua.es}
%\and
%IEEEauthorblockN{4\textsuperscript{th} Alberto~Garcia-Garcia}
%\IEEEauthorblockA{\textit{Institute of Space Sciences (ICE-CSIC)} \\
%\textit{Spanish National Research Council (CSIC)}\\
%Spain \\
%garciagarcia@ice.csic.es}
%\and
%\IEEEauthorblockN{5\textsuperscript{th} Sergio~Orts-Escolano}
%\IEEEauthorblockA{\textit{Dep. of Computer Science and AI} \\
%\textit{University of Alicante}\\
%Alicante, Spain \\
%sorts@ua.es}
%\and
%\IEEEauthorblockN{6\textsuperscript{th} Jose~Garcia-Rodriguez}
%\IEEEauthorblockA{\textit{Department of Computer Technology} \\
%\textit{University of Alicante}\\
%Alicante, Spain \\
%jgarcia@dtic.ua.es}
%\and
%\IEEEauthorblockN{6\textsuperscript{th} Markus~Vincze}
%\IEEEauthorblockA{\textit{Vision for Robotics Laboratory} \\
%\textit{TU Wien}\\
%Wien, Austria \\
%vincze@acin.tuwien.ac.at}
%}

\maketitle

\begin{abstract}
Synthetic data generation has become essential in last years for feeding data-driven algorithms, which surpassed traditional techniques performance in almost every computer vision problem. Gathering and labelling the amount of data needed for these data-hungry models in the real world may become unfeasible and error-prone, while synthetic data give us the possibility of generating huge amounts of data with pixel-perfect annotations. However, most synthetic datasets lack from enough realism in their rendered images. In that context UnrealROX generation tool was presented in 2019, allowing to generate highly realistic data, at high resolutions and framerates, with an efficient pipeline based on Unreal Engine, a cutting-edge videogame engine. UnrealROX enabled robotic vision researchers to generate realistic and visually plausible data with full ground truth for a wide variety of problems such as class and instance semantic segmentation, object detection, depth estimation, visual grasping, and navigation. Nevertheless, its workflow was very tied to generate image sequences from a robotic on-board camera, making hard to generate data for other purposes. In this work, we present UnrealROX+, an improved version of UnrealROX where its decoupled and easy-to-use data acquisition system allows to quickly design and generate data in a much more flexible and customizable way. Moreover, it is packaged as an Unreal plug-in, which makes it more comfortable to use with already existing Unreal projects, and it also includes new features such as generating albedo or a \textit{Python API} for interacting with the virtual environment from \textit{Deep Learning} frameworks.
\end{abstract}


\input{sections/introduction}
\input{sections/related_works}
\input{sections/system}
\input{sections/applications}
\input{sections/experiments}

\input{sections/conclusion}
\input{sections/future}

\section*{Acknowledgment}
This work has been funded by the Spanish Government PID2019-104818RB-I00 grant for the MoDeaAS project, supported with Feder funds. This work has also been supported by Spanish national grants for PhD studies FPU17/00166, ACIF/2018/197 and UAFPU2019-13. Experiments were made possible by a generous hardware donation from NVIDIA.

%% EXAMPLES

%% Figures
%\begin{figure}[htbp]
%\centerline{\includegraphics{figures/fig1.png}}
%\caption{Example of a figure caption.}
%\label{fig}
%\end{figure}

%% Tables
% \begin{table}[htbp]
% \caption{Table Type Styles}
% \begin{center}
% \begin{tabular}{|c|c|c|c|}
% \hline
% \textbf{Table}&\multicolumn{3}{|c|}{\textbf{Table Column Head}} \\
% \cline{2-4} 
% \textbf{Head} & \textbf{\textit{Table column subhead}}& \textbf{\textit{Subhead}}& \textbf{\textit{Subhead}} \\
% \hline
% copy& More table copy$^{\mathrm{a}}$& &  \\
% \hline
% \multicolumn{4}{l}{$^{\mathrm{a}}$Sample of a Table footnote.}
% \end{tabular}
% \label{tab1}
% \end{center}
% \end{table}

\bibliographystyle{IEEEtran}
\bibliography{references}

\vspace{12pt}

\end{document}

%% file: acronyms.tex
\newacronym{ue4}{UE4}{Unreal Engine 4}
\newacronym{fps}{FPS}{frames per second}
\newacronym{api}{API}{Application Program Interface}
\newacronym{CHALET}{CHALET}{Cornell House Agent Learning Environment}
\newacronym{HoME}{HoME}{Household Multimodal Environment}
\newacronym{THOR}{THOR}{THe House of inteRactions}
\newacronym{MINOS}{MINOS}{Multimodal Indoor Simulator}
\newacronym{URDF}{URDF}{Unified Robot Description File}

%% file: sections/introduction.tex
\section{Introduction}
\label{sec:introduction}

In the past few years, the evolution of vision-based deep learning architectures have confirmed its support on synthetically generated data in order to save the lack of huge amounts of labeled image data. It is well known that gathering relevant amounts of images with ground truth in the real world is an expensive and tedious task, if not impossible in some cases. Virtual environments with powerful rendering technologies and engines allow generating huge amounts of synthetic images that can be automatically labeled.

In 2018, the Robotrix dataset\cite{Garcia2018robotrix} was presented as a synthetically generated multi-purpose dataset for robotics. The main contribution of this dataset was the quality and variety of the data that it presented, including RGB images, depth and normal maps, and instance and semantic segmentation masks at 1080p resolution and 60 \gls{fps}. The tool that was developed to generate it, UnrealROX \cite{Martinez2019unrealrox}, is based on \gls{ue4}, one of the most widespread video game engines and, therefore, it is one of the most cutting-edge and evolving platforms for realistic 3D rendering. The tool was developed specifically for generating that robotic-oriented dataset, and due to that, it was not engineered for getting data in some other ways that could be useful for different purposes, such us reinforcement learning, or simply for random or scripted data gathering. In this way, in order to make UnrealROX a truly data generator for a wider range of applications, we decoupled the main functionalities from it, so that it would be easy to isolate the data acquisition tool, and to add any custom behaviour that any developer could need for generating its own data. In this paper, we present this new and more flexible workflow, and all the new functionalities that we added to the tool, as well as several qualitative experiments that prove the usefulness of synthetic data from Unreal Engine in a variety of vision-based deep learning architectures.

This paper is organized as follows. First, Section \ref{sec:related_works} analyzes already existing environments for synthetic data generation and puts our proposal in context. Next, Section \ref{sec:system} describes new features on the system that make it a flexible tool for generating synthetic data from \gls{ue4} environments. Section \ref{sec:applications} presents different areas inside machine learning where data generated through this tool could be useful, and Section \ref{sec:experiments} includes a set of qualitative experiments carried out for testing the tool usefulness. At last, Section \ref{sec:conclusion} summarizes the paper and draws conclusions about this work, and Section \ref{sec:future} points out the limitations and possible future extensions for the tool.

%% file: sections/related_works.tex
\section{Related Works}
\label{sec:related_works}

Synthetic environments have been used for a long time to benchmark vision and robotic algorithms \cite{Butler2012}. Recently, their importance has been highlighted for training and evaluating machine learning models, not only for robotic vision problems \cite{Brodeur2017} \cite{Ros2016} \cite{Mahler2017}, but many others \cite{Mu2020animals} \cite{Wang2019crowdcounting} \cite{Richardson2016face} \cite{Garnett2020lanedetection} \cite{Ludl2020humanpose}. Due to the increasing need for samples to train such data-driven architectures, there exists an increasing number of synthetic datasets.

In the context of generating data for indoor robotic tasks (the original proposal of UnrealROX), we already reviewed \gls{CHALET} \cite{Yan2018}, a 3D house simulator for manipulation and navigation learning built in Unity 3D, \gls{HoME} \cite{Brodeur2017}, a multimodal household environment for AI learning from visual, auditive, and physical information within realistic synthetic environments sourced from SUNCG, AI2-\gls{THOR} \cite{Kolve2017}, a framework for visual AI research which consists of near-photorealistic synthetic 3D indoor scenes in which agents can navigate and change the state of actionable objects, and \gls{MINOS} \cite{Savva2017}, a simulator for navigation in complex indoor environments. Their main shortcomings range from lack of realism, full 3D robot meshes, or first, third or multi camera support.
Few more recent generators or environments are VirtualHome, Habitat or ElderSim.

\vspace*{0.1cm}\noindent \textbf{VirtualHome} \cite{Puig2018virtualhome} is a multi-agent platform to simulate activities in a household. Agents are represented as humanoid avatars, which can move and interact with the environment through high-level instructions. It can be used to render videos of human activities, or train agents to perform complex tasks. Its strongest point is being able to generate whole sequences from just some high-level instructions, and its weakest one is probably the render and animation realism. %Nevertheless, what it offers is pretty impressive.

\vspace*{0.1cm}\noindent \textbf{Habitat} \cite{Savva2019habitat} enables the training of embodied agents (virtual robots) in a highly efficient photorealistic 3D simulation. It uses its own fast and optimized 3D simulator, and also offers an \gls{api} for end-to-end development of embodied AI algorithms: defining tasks (e.g. navigation, instruction following, question answering), configuring, training, and benchmarking embodied agents.

\vspace*{0.1cm}\noindent \textbf{ElderSim} \cite{Hwang2020eldersim} is a synthetic action simulation platform that can generate synthetic data on elders' daily activities. It can generate realistic motions of synthetic characters for 55 kinds of activities, with several customizable data-generating options and output modalities. It is based on Unreal Engine, and provides an user interface for making data generation easy. ElderSim is focused on generating data from multiple and fixed points of view, and oriented to action and human pose detection.

When presenting \textit{UnrealROX}, we already stated that few already existent tools and environments served as inspirations for the project. They were UnrealCV, Gazebo, and NVIDIA's Isaac Sim. UnrealCV \cite{Qiu2016}\cite{Qiu2017} is a project that extends \gls{ue4} to create virtual worlds and ease communication with computer vision applications. UnrealCV consists of two parts: server and client. The server is a plugin that runs embedded into an \gls{ue4} game. We took the main concept and design behind UnrealCV and implemented the whole pipeline inside \gls{ue4} itself to be more efficient and customizable. Gazebo \footnote{http://http://gazebosim.org/}, on the other hand, is a well-known robot simulator that enables accurate and efficient simulation of robots in indoor and outdoor environments. It integrates a robust physics engine (Bullet, ODE, Simbody, and DART), advanced 3D graphics (using OGRE), and sensors and noise modelling. Lastly, NVIDIA's Isaac Sim\footnote{https://developer.nvidia.com/isaac-sim} is a virtual simulator for robotics that lets developers train and test their software using highly realistic virtual simulation environments. When UnrealROX was presented, this project was in an early development phase, and ran under \gls{ue4}. Currently, the project has been integrated with \textit{NVIDIA Omniverse}, an open platform built for virtual collaboration and real-time photorealistic simulation.

\subsection{Our Proposal in Context}

UnrealROX focused on simulating a wide range of common indoor robot actions, both in terms of poses and object interactions, by leveraging a human operator to generate plausible trajectories and grasps in virtual reality. Now, we propose to take advantage of all this data-generation potential, making it easier to use it for a wider variety of goals. Unreal Engine is a widespread platform in continuous development, and any new hardware, such as a virtual or augmented reality headsets, motion capture devices, or similar, will presumably provide support for it. So, we find highly interesting to have the possibility of generating any kind of data, in a fast, easy and customizable way, with this versatile and powerful engine.

Along with the generation of raw data (RGB-D/3D/Stereo) and ground truth (2D/3D class and instance segmentation, 6D poses, and 2D/3D bounding boxes), we now provide albedo, shading maps, interaction information, and a Python \gls{api}. Moreover, the pose of virtual agents with skeleton (human meshes, robots or hands, for example) can also be projected over RGB or mask images, since joint 6D pose is provided in a frame-by-frame basis.

%Everything with a renewed and flexible system for obtaining data, and improved strategies for acquiring some of them, such as segmentation masks.

%% file: sections/system.tex
\section{System}
\label{sec:system}

\begin{figure*}[htbp]
\centerline{\includegraphics[width=0.85\textwidth]{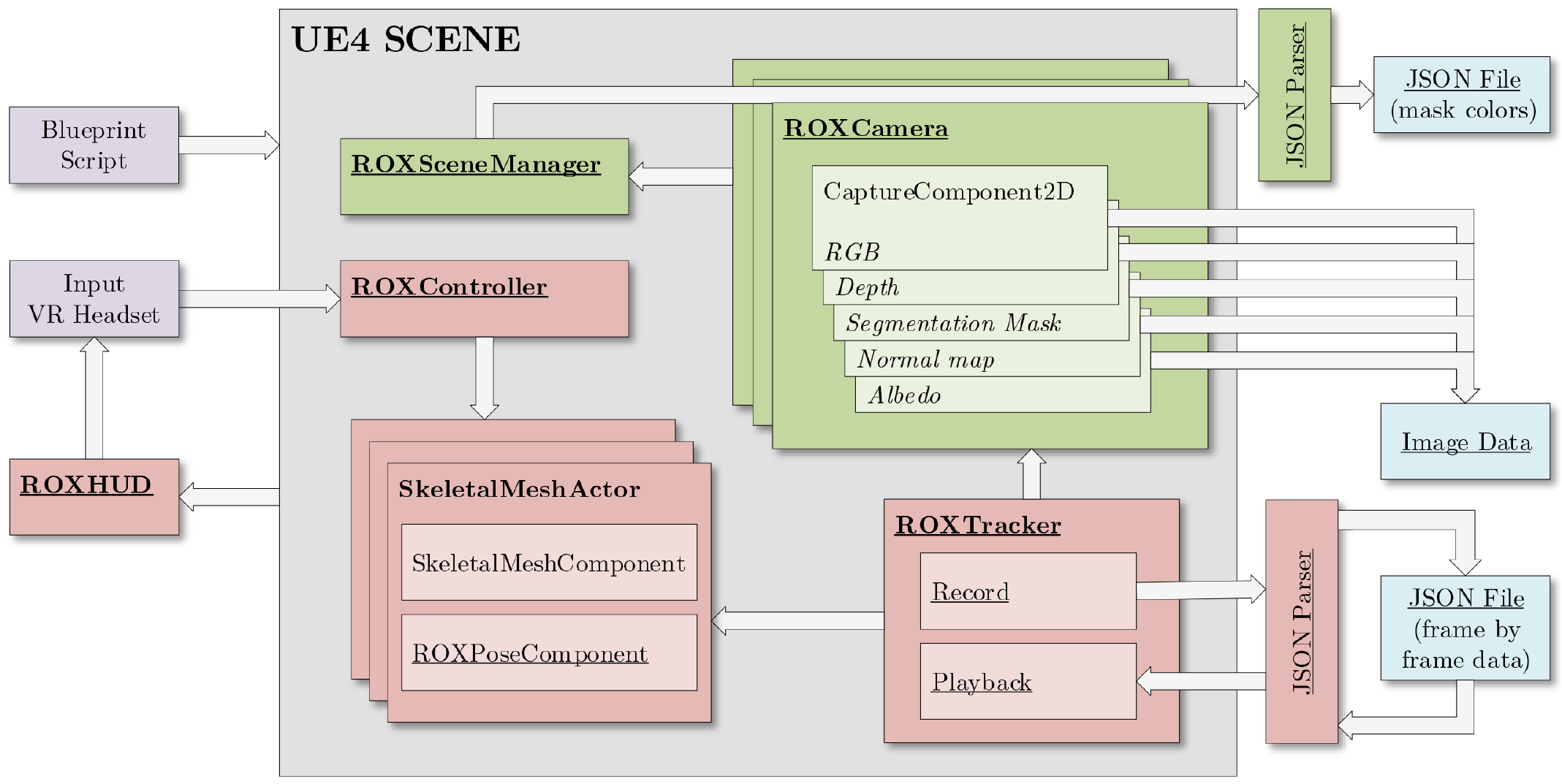}}
\caption{UnrealROX+ system diagram. In green, decoupled data acquiring system that can be used with more flexibility. In red, adapted logic from UnrealROX for using new decoupled data acquiring system, not essential for generating data. Custom classes (actors or components) are underlined.}
\label{fig:UROXPlusdiagram}
\end{figure*}

\begin{figure}[htbp]
\centerline{\includegraphics[width=8cm]{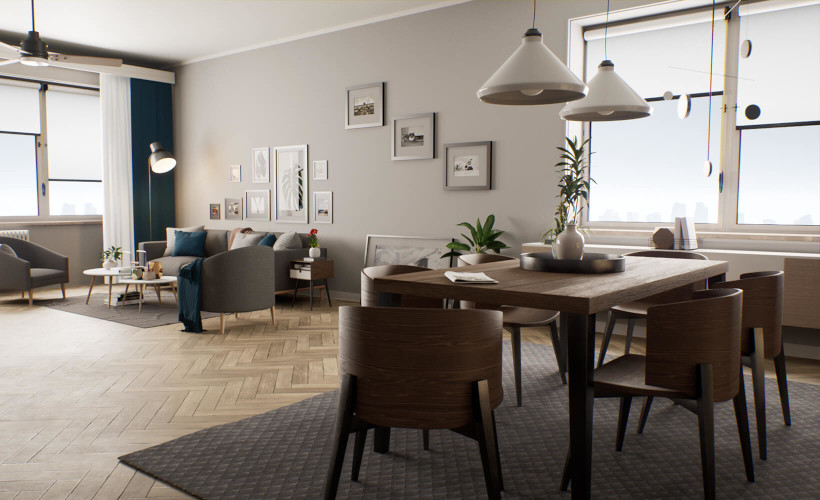}}
\caption{Real time ray-tracing snapshot on \gls{ue4} from the Archviz Interior Rendering sample project.}
\label{fig:raytracing}
\end{figure}

As we already stated when we first presented our tool in \cite{Martinez2019unrealrox}, the rendering engine we chose to generate photorealistic RGB images was \gls{ue4}. The reasons for this choice are the following ones: (1) it is arguably one of the best game engines able to produce extremely realistic renderings, (2) beyond gaming, it has become widely adopted by Virtual Reality developers and indoor/architectural visualization experts so a whole lot of tools, examples, documentation, and assets are available; (3) due to its impact across various communities, many hardware solutions offer plugins for \gls{ue4} that make them work out-of-the-box; and (4) Epic Games provides the full C++ source code and updates to it, so the full suite can be used and easily modified for free. Arguably, the most attractive feature of \gls{ue4} that made us take that decision is its capability to render extremely photorealistic imagery of synthetic environments. Some \gls{ue4} features that enable this realism are: physically-based materials, pre-calculated bounce light via Lightmass, stationary lights, post-processing, and reflections. In addition, real-time ray tracing rendering was introduced as beta feature in Unreal Engine's 4.22 version on April 2019 \cite{UnrealRayTracing2019}, and nowadays it is completely available \cite{NvidiaRayTracing2020}. This allowed us to generate ray-tracing-rendered images, as the one showed in Figure \ref{fig:raytracing}, both in real time and offline, without modifying the tool.
%The continuous developing of the engine is one of its greatest advantages.

UnrealROX was initially developed as a tool for generating synthetic data from virtual photorealistic 3D environments, as we already know. However, its workflow was totally focused on immersing a human agent in these environments through virtual reality, \textit{recording} all its movements and interactions, and later \textit{rebuilding} the sequence frame by frame in order to have the time needed to retrieve all kind of image data, at high frame rates, high resolutions, and high realism. After releasing RobotriX \cite{Garcia2018robotrix} and UnrealROX, we continued using the tool for generating data for different kinds of problems, and several modifications were introduced to adapt its workflow. We end up deciding to decouple the data acquisition feature of the tool from the main workflow that was used for generating RobotriX (\textit{record}, \textit{rebuild} and \textit{acquire}). In this way, the tool now allows to acquire image and other data from the virtual environment very easily, directly from the \gls{ue4} visual scripting language, which is also known as \textit{blueprints}. That means that preparing scripts for generating images of objects from different point of views, or moving a camera (randomly or not) across the scene is truly fast. We also kept the possibility of generating data with the original workflow, but adapting the code for calling the new decoupled data acquiring system.

The other big improvement for the tool was packaging it as a Unreal Engine plug-in, which is remarkable because it allows to incorporate all these interesting features to any existing project very easily. These two main changes in the tool implied many other little ones that we are going to review in this chapter, as well as other additions that the tool received during this period. For example, along with RGB, depth maps, normal maps, and instance segmentation masks, we also provide albedo directly retrieved from \gls{ue4}, and its corresponding shading maps that can be post processed later. We can see the new system diagram for UnrealROX+ in Figure \ref{fig:UROXPlusdiagram}. The tool itself, as well as more extensive documentation is available as open-source software\footnote{https://github.com/3dperceptionlab/unrealrox-plus}.

\subsection{Data acquiring subsystem}

The motivation for this tool revision was decoupling the ground-truth image acquisition in order to make it faster and more flexible to use for any scenario. We decided to encapsulate all these functionalities under a custom class inheriting from the basic Unreal class \textit{Camera}, that we called \textit{ROXCamera}. In this way, each \textit{ROXCamera} in the scene has the possibility to be configured and requested to retrieve some concrete data, separately. Automating this process for more than one camera is as simple as keeping a data structure with their references and using them when needed. Moreover, these cameras can be referenced and called from Blueprints very easily.

\begin{figure*}[htbp]
    \centering
    \includegraphics[width=0.32\textwidth]{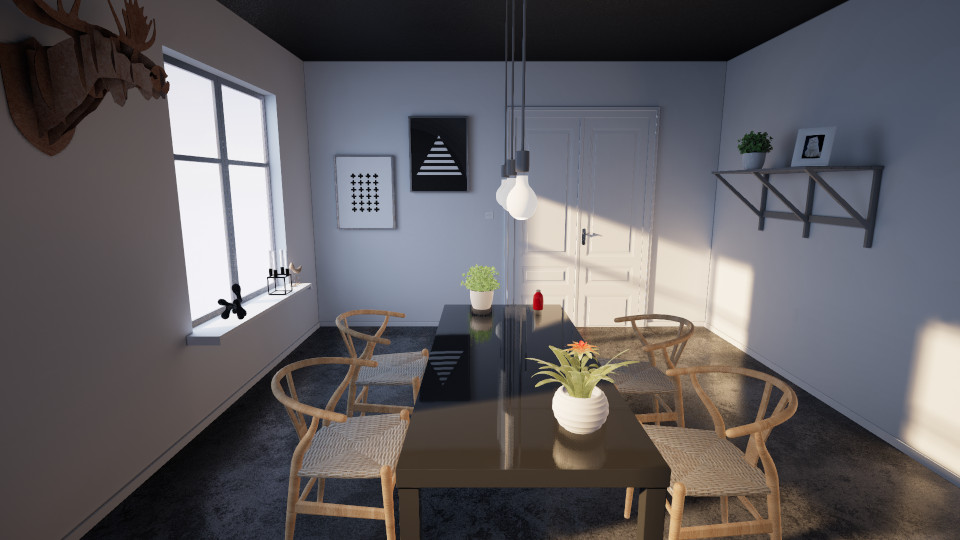}
    \includegraphics[width=0.32\textwidth]{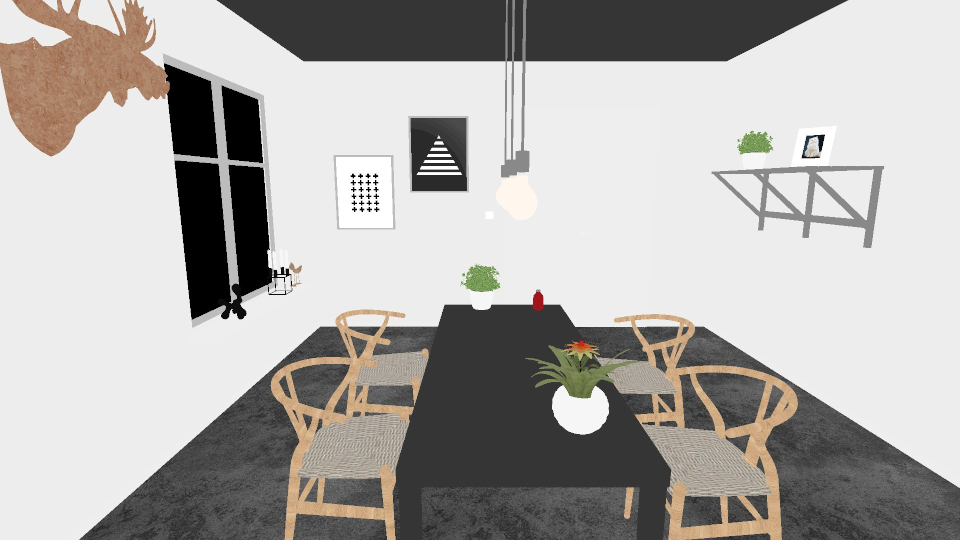}
    \includegraphics[width=0.32\textwidth]{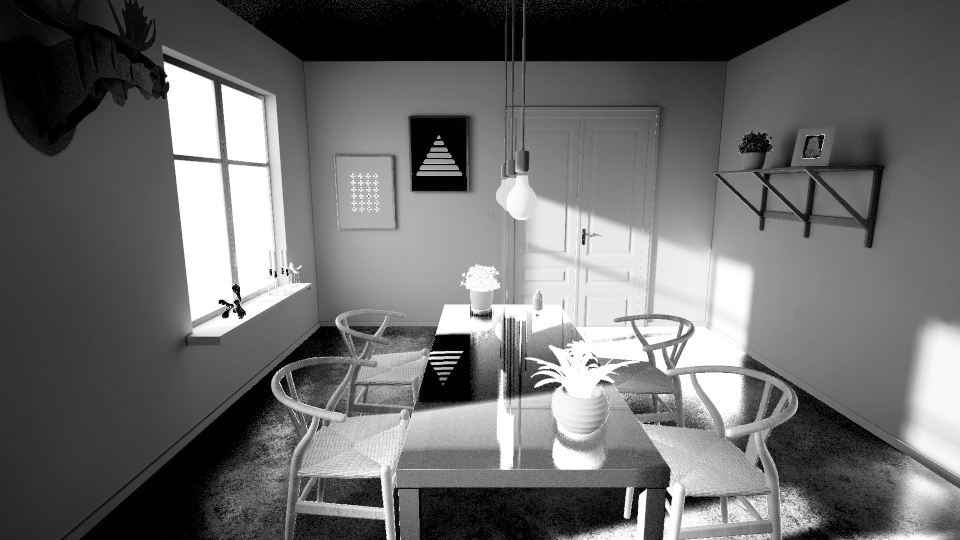}
    \includegraphics[width=0.32\textwidth]{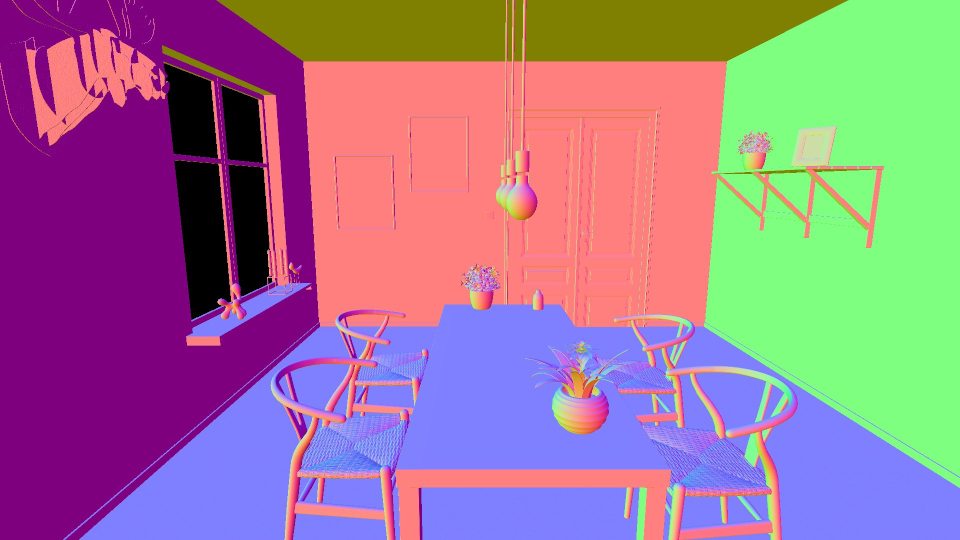}
    \includegraphics[width=0.32\textwidth]{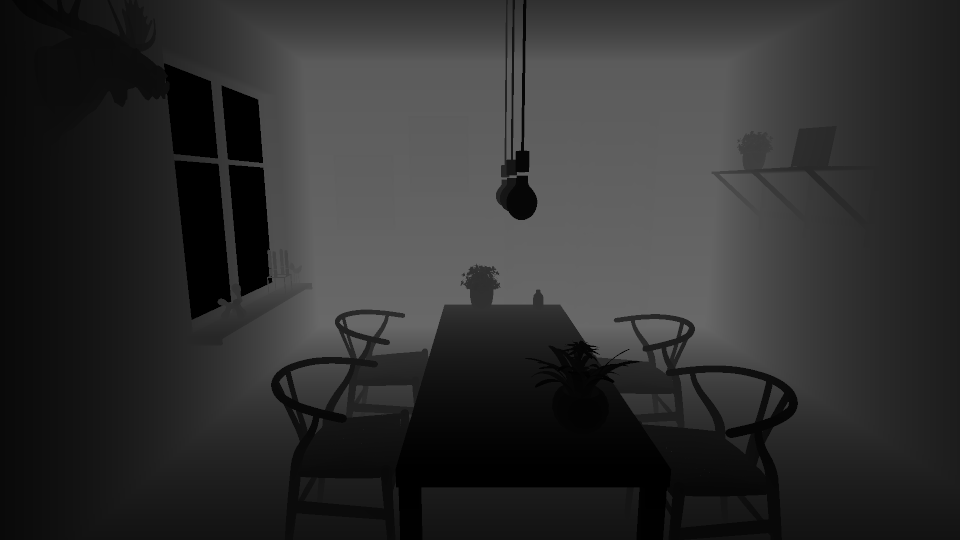}
    \includegraphics[width=0.32\textwidth]{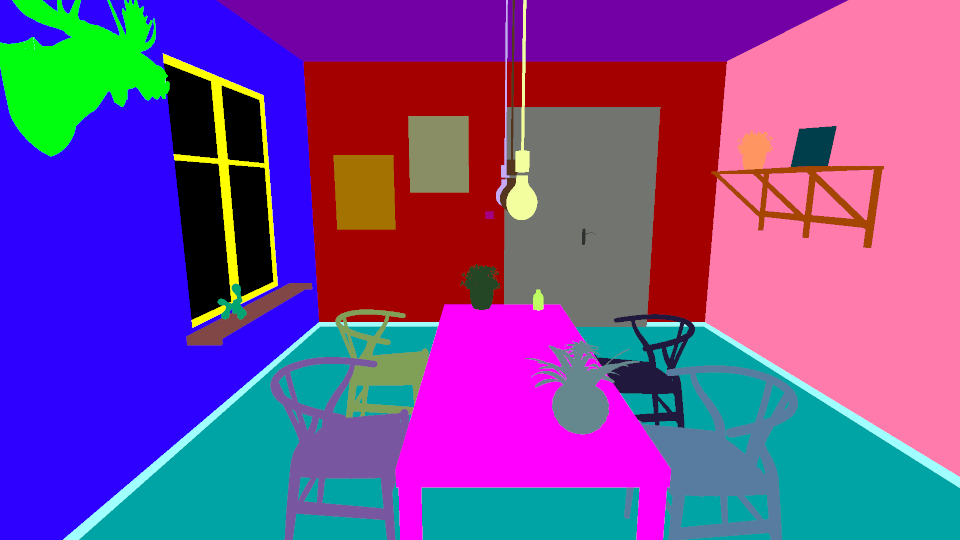}
    \caption{UnrealROX+ image data. From left to right and top down: RGB, albedo, shading map, normal map, depth map and instance segmentation map. Shading map is not generated directly from \gls{ue4}, it is computed later from RGB and albedo.}
\label{fig:UROXData}
\end{figure*}

This encapsulation simplifies usage, but we also reinterpreted the data-acquiring logic itself to make it more elegant. Original UnrealROX retrieved all image data directly from the main viewport, which is the point of view that is being rendered in \gls{ue4} user's screen. Using main viewport when generating data from several cameras implies continuously switching the viewport source camera, as well as the rendering mode of the whole scene (for generating RGB, depth, normal maps, etc). This strategy was rough, especially knowing that \gls{ue4} provides a dedicated and flexible entity for capturing image data from the scene without having to use a standard camera and the viewport: the \textit{SceneCapture2D} actor, or more precisely, the \textit{CaptureComponent2D}, which is the component that contains its functionality.
%It can look the same as a normal camera, but it is designed to be pretty flexible.
It can be used for rendering (with its own custom render mode) a concrete point of view to a \textit{RenderTarget}, instead of using the viewport.
%So, that point of view and render mode are treated separately.
\textit{RenderTarget} is another entity, which encapsulates the data that is rendered by \textit{CaptureComponent2D} and can dump it to a \textit{Texture}, or to a file, for example.

Our \textit{ROXCamera} class is designed to create one \textit{CaptureComponent2D} entity for each type of data that must be generated from that point of view, and everything is configured automatically. As we have one \textit{CaptureComponent2D} for each render mode, we don't have to worry about changing the general render mode. The file saving logic can be easily managed from the \textit{RenderTarget}, where we can select the color codification or the file format (which differs among different image data).

\subsection{Segmentation masks}

Semantic segmentation masks are one of the most expensive data to provide in any real dataset. For any real-world image, manually discriminating at pixel level for generating a reliable mask (for semantic or instance segmentation) is a very tedious task. So, obtaining this kind of information in a pixel-perfect and automatic way is one of the most relevant reasons to use synthetic generators for creating huge image datasets.

In our first version of UnrealROX, inspired by a preliminary implementation of UnrealCV \cite{Qiu2016}, we modified the engine's source code (simply changing a flag) in order to be able to modify at will the vertex color as a post process operation, and thus being able to render and capture these kind of images. The fact of recompiling the whole engine shouldn't be a problem, since \gls{ue4} is open source, but both the development and the use by other teams become harder and slower in this way, so we developed alternatives for generating these data. At the beginning of each generating execution, every mesh on the scene is gathered, and a reference to it is stored. Then, we developed two approaches:

\begin{itemize}
    \item A plain-color material instance is automatically created and associated with each of those meshes. A distinct color is assigned to each mesh, in such way that, when rendering the scene in \textit{base color} mode (i.e. \textit{unlit}, without lighting computations) with these new materials applied, we get pixel-perfect segmentation masks. Obviously, we have to keep a data structure with references to both original and plain-color materials in order to be able to swap between these two rendering modes at any time.
    
    \item A \textit{post process material} is applied depending on the value stored at the \textit{custom stencil buffer}. This is a special GPU buffer which we have the possibility to modify. For each mesh in the scene, we can assign a custom stencil value, resulting in the buffer storing that value for each pixel that belongs to that mesh. Then, it can be used in a post process material, which act as shaders in \gls{ue4}, performing operations at pixel level. Post process materials can be applied directly to a \textit{CaptureComponent2D} entity, which is great for our workflow.
\end{itemize}

Acquiring and storing these mask images is a delicate task, since their stored pixel values must match exactly the ones that we set numerically in the engine. In order to avoid little color variances inside the same region, tone mapping is disabled (or at least it is applied before the post process), and linear color read from \textit{RenderTarget} must be transformed to \textit{sRGB} color space with gamma equal to 1.

It is worth mentioning that the plain-color material approach has a remarkable disadvantage. Swapping materials for every mesh on the scene is way more computationally demanding than the \textit{render mode} switch that we had in the first UnrealROX version, or the post process material alternative. Nevertheless, we still consider that it is viable option because we can still generate segmentation masks separately from the rest of ground truth information (as long as we are generating data offline). In this way, we will only have to perform the material swap once at the beginning of the generation, and once more at the end. On the other hand, the post process material approach has the limitation of just being able to codify stencil 256 different values, so it only should be used if the the scene has less than that number of different meshes.

\subsection{Reflectance and Shading maps}

\textbf{Reflectance} is another image data which is interesting for several problems, such as Intrinsic Image Decomposition \cite{bell2014intrinsic} or Inverse rendering \cite{yu2019inverserendernet}. It represents the inherent color from surfaces, the color they naturally reflect, without any lightning influence or computation (shadows, specularities, etc). 
%So, it gives us the true color information, without lights or shadows.
In computer graphics research area is common to use the term \textit{albedo}, which technically represents the percentage of light that is reflected by a surface \cite{coakley2003albedo}, but can be used as well as \textit{reflectance} for referring this kind of images with only the diffuse reflection component from an illumination model \cite{sato1997reflectance}. We will use reflectance or albedo indistinctly.
This data can be directly obtained in \gls{ue4} through the render mode \textit{base color}, that can be chosen in \textit{CaptureComponent2D} structure. As we already verified when working on generating segmentation masks, pixel color information obtained through images captured in base color are perfectly accurate when retrieved as described.

\vspace*{0.1cm}\noindent \textbf{Shading maps}, on the other side, provide us information about lightning for each pixel, mainly how much it is darkened or lightened starting from the base color provided by reflectance. Image decomposition is defined as follows \cite{bell2014intrinsic}:

$$  I = R \odot S $$

being \textit{I} the composed image, \textit{R} the reflectance layer, and \textit{S} the shading layer. So \textit{I} is defined as the element-wise multiplication ($\odot$) between reflectance and shading. In this case, we don't acquire this data directly from the engine. Instead, we compute it from RGB and albedo images. Starting from the equation defined previously, we can infer that:

%\textbf{Shading maps}, on the other side, represent the amount of light that is present in each pixel. 

$$  S = I \varoslash (R + \epsilon) $$

which means that shading map can be computed as the matrix element-wise division ($\varoslash$) between the image and its correspondent albedo. Since a division must be performed, we can avoid dividing by zero adding a very little value ($\epsilon$) to albedo pixel values. For the vast majority of scenarios, white lighting can be assumed, so shading maps are widely represented as gray scale. That means that light affects to all three RGB channels equally. An example of albedo and shading map images, along with its correspondent RGB and other images, can be seen at Figure \ref{fig:UROXData}.

\subsection{Skeletal Meshes}

One of the key parts in the development of the original UnrealROX was storing every joint (bone) 6D transformation of the controlled pawn, which is essentially an skeleton (i.e. \textit{SkeletalMeshActor} in \gls{ue4}). We needed to keep all that information frame by frame in order to later be able to retrieve the exact pose of the skeleton for acquiring data offline. This goal was reached by creating our own class (\textit{ROXBasePawn}) that inherits from the basic \textit{Pawn} class, and adding to it all the logic needed to save and later recover joint transformations. However, this approach had a huge disadvantage: it was mandatory to use our \textit{ROXBasePawn} class in order to be able to use the offline data-acquiring workflow with an skeletal mesh. This may imply migrate code and assets from an already existing class, to ours. The alternative we developed to better align this with \gls{ue4} programming standards and the fact of delivering a plug-in, consisted of encapsulating this joint-retrieving logic as a component, so that it can be added to any existing actor with a \textit{SkeletalMeshComponent}.
%So, that component encapsulates this little logic for storing and recovering joint transformation in a much flexible way, and it is definitely a preferred one as we are packaging all this as a plug-in.

\subsection{Python communication for Reinforcement Learning}

One of the biggest future works mentioned on the original UnrealROX paper was a \textit{Python \gls{api}} in order to be able to modify Unreal Engine's scene from deep learning frameworks. This workflow is specially useful for reinforcement learning models, since they are systems that successively improve the actions they take in an environment by maximizing a reward function. So, these models need to modify the scene, see what happens (retrieving data from it) and check how the reward function behaves. In order to make our tool useful for these applications, and therefore, make \gls{ue4} usable for reinforcement learning, we developed several commands that can be thrown from Python programming language, and caught by Unreal Engine, performing the correspondent actions in the virtual scene. The system is defined as a client-server architecture, where Python API acts as client, and our \gls{ue4} plug-in as server, and it works both locally and remotely. We defined the following groups of commands (further documentation is available at GitHub repository):

\begin{itemize}
    \item \textbf{Lists}: In order to interact with actors and meshes in the scene, we must know which they are, so we implemented commands to list them depending of their type, such as \texttt{actor\_list}, \texttt{object\_list}, \texttt{camera\_list}, \texttt{skeletal\_list}, among others.
    
    \item \textbf{Transformations}: To apply transformations over actors in the scene we have \texttt{move}, \texttt{rotate} and \texttt{scale}, as well as their correspondent commands for retrieving the current location, rotation and scale.
    
    \item \textbf{Cameras}: We can, for example, spawn and orientate cameras with \texttt{spawn\_camera} and \texttt{camera\_look\_at}, respectively.
    
    \item \textbf{Data acquiring}: We can acquire image data from a concrete camera with \texttt{get\_rgb}, \texttt{get\_depth}, \texttt{get\_normal}, \texttt{get\_instance\_mask} and \texttt{get\_albedo}. Other commands can configure this operations, and also acquire other kind of data, such as bounding boxes with \texttt{get\_3d\_bounding\_box}.
\end{itemize}

\subsection{Grasping subsystem and interaction information}

The grasping subsystem that we used for the robot-object interactions in RobotriX, which was included on the original version of UnrealROX, was completely decoupled and evolved on its own project with \textit{UnrealGrasp} \cite{Oprea2019unrealgrasp}, which is still being developed in parallel. It will also be released as a plug-in, completely compatible with UnrealROX+. This also decoupled interaction information that was added to the rest of frame-by-frame data stored for movable \textit{StaticMeshActor} actors. Data gathered for each frame was: if that object was being touched (overlapped), grasped, or none, by the controlled actor. We left this information to be given separately by UnrealGrasp, so UnrealROX+ just provide a list of overlapping actors. This is, for each actor, the ones that are being overlapped in that concrete frame, not distinguishing static from skeletal meshes.

%\subsubsection{Contact points extraction}

%Additionally, we were working on generating some kind of heat map showing the contact points that a hand (or a robot clamp) are most likely to use for getting a stable grasping. This data that can help a robotic system to identify a suitable grasp given an object. These heat maps are generated after repeating several times suitable grasps over the same object.\\

%In the same way as previous interaction information, we developed this to work with the grasping system that we decoupled, so it was left out from the base code of UnrealROX+. We developed it using a material in the interacted object that detects and projects on itself the overlapping section between the ``grasp'' actor and the own object. Then, this material is exported as a texture, and stored as an image.

%% file: sections/applications.tex
\section{Applications}
\label{sec:applications}
UnrealROX+ provides a great variety of data. The set of tasks and problems that can be addressed using such data ranges from low to high-level ones. Some of the most relevant low-level tasks include:

\begin{itemize}
    \item Depth Estimation: Both monocular (deep learning models estimating depth from RGB) and stereo (estimate 3D information from a pair of displaced cameras) are enabled by our system. Stereo may be reached manually by placing two cameras, but we implemented it natively so that it was more comfortable to work with camera pairs.
    \item Object detection and pose estimation: We provide rich information about objects in the scene (instance and class segmentation masks, 2D and 3D bounding boxes, 6D position) to work in these problems.
    \item Instance/Class segmentation: Instance masks are directly provided, and they can be post-processed for enabling class segmentation as well.
    \item Normal estimation: Estimating the normal map of a given surface is an important previous step for many other tasks. For instance, certain algorithms require normal information in a point cloud to extract possible grasping points. UnrealROX+ provides per-pixel normal information.
    \item Intrinsic image decomposition: Image decomposition in its intrinsic parts, reflectance and shading.
\end{itemize}

That low-level data enables other higher-level tasks that either make use of the output of those systems, or take the low-level data as input, or both:

\begin{itemize}    
    \item Hand pose estimation: UnrealROX+ provides the 6D pose of every skeleton joint, so hand pose can be estimated. It is useful for gesture detection, for example.
    \item Human pose estimation: Again, 6D pose of every skeleton joint is provided, so skeleton pose estimation can be trained with data generated by our tool.
    \item Obstacle avoidance and navigation: By leveraging various types of low-level information such as RGB images, depth maps, bounding boxes, and semantic segmentation, robots can learn to avoid obstacles (by detecting objects and estimating their distance) and even navigate in indoor environments.% (by building a map to localize themselves in the indoor scene while avoiding objects and walls, and being able to reason semantically to move intelligently).
\end{itemize}

Data generated by UnrealROX+ can be useful for other applications, although in this case they may need further development or combination with other tools:
\begin{itemize}
    \item Reinforcement learning: Python \gls{api} enables using the tool for generating data on demand directly from a reinforcement learning model in training process.
    \item Visual grasping: In combination with \textit{UnrealGrasp}, agents in the virtual scene can perform plausible grasps and record RGB and hand pose data.
    %\item Grasping points estimation: Also helped by UnrealGrasp, a heat map with statistical information about feasible grasping points may be generated.
    \item Action detection and prediction: RGB and skeleton pose data are provided directly, and further labelling for actions at multiple granularity levels can be made afterwards with proper tools. Even some kind of automatic labelling can be developed if pre-designed animations are being used for performing those actions.
    
\end{itemize}

%% file: sections/experiments.tex
\section{Experiments}
\label{sec:experiments}
In the previous section we showed multiple potential applications to which our data generator and the corresponding ground truth could be applied to train machine learning systems. Besides tasks such us depth estimation and visual grasping, which relied on data generated via the previous workflow \cite{Martinez2019unrealrox}, we selected two more applications in which the new UnrealROX+ is exploited.
We show the usefulness of the generated data in the aforementioned tasks, complementing the already available datasets from the real domain.

%We will see how the limited a of fully labelled datasets is mitigated using our synthetically generated data.

%In this section,  There, UnrealROX+ new workflow allowed to generate data that addressed a lack of available labelled data, or simply complemented it. Usefulness on other applications, such as depth estimation \cite{Martinez2019unrealrox} or visual grasping \cite{Oprea2019unrealgrasp}, was already treated in previous papers with the original workflow.

%to experiment with them in order to prove the effectiveness of our improved data acquiring tool. Those two representative problems are: Hand joint (3D) position estimation and 6D object pose estimation.

\subsection{Hand joint position estimation}

\begin{figure}
    \centering
    \includegraphics[width=0.49\linewidth]{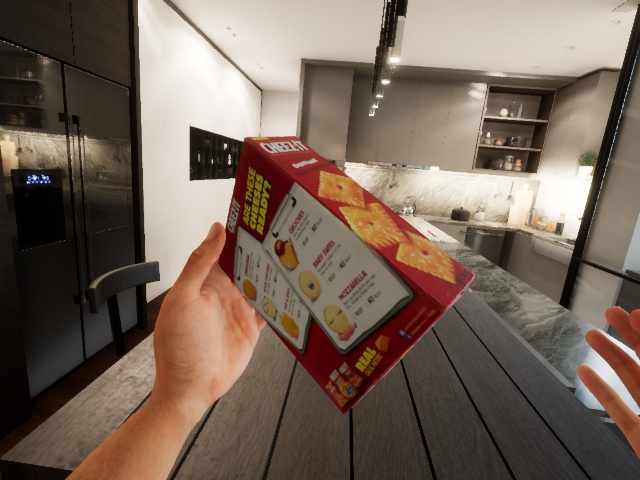}
    \includegraphics[width=0.49\linewidth]{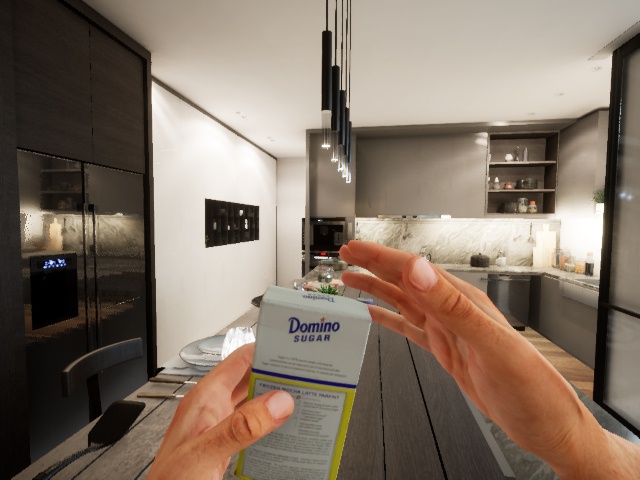}
    \includegraphics[width=0.49\linewidth]{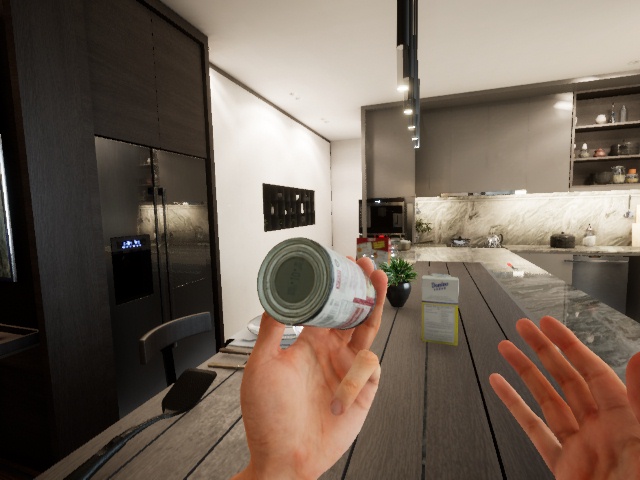}
    \includegraphics[width=0.49\linewidth]{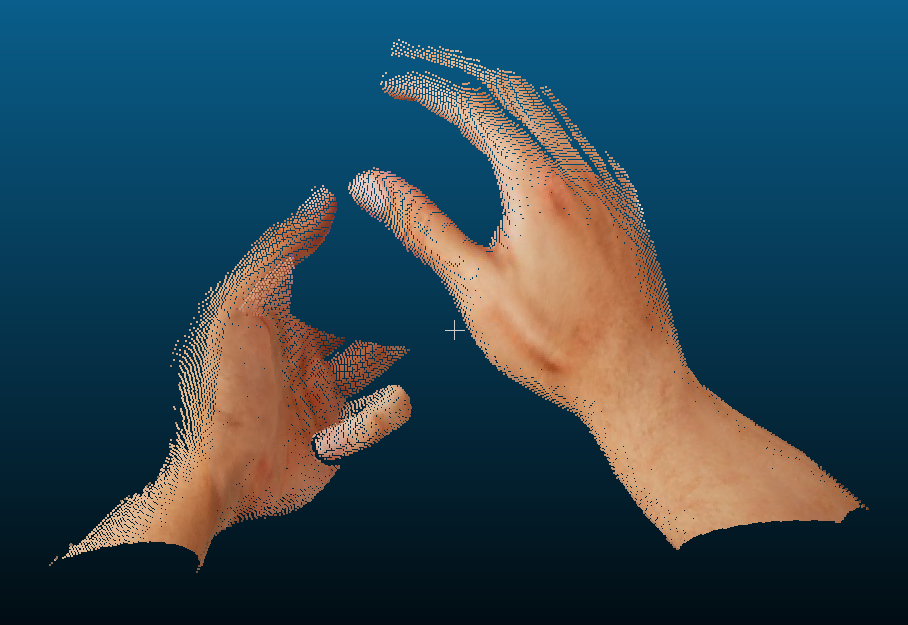}
    
    \caption{This figure shows a subset of samples images with 3 of the objects available on the UnrealROX-generated dataset presented in \cite{castro2020grasphands}, as well as a point-cloud example. Point clouds are computed from the dataset and used for feeding GraphHands.}
    \label{fig:grasp_dataset}
\end{figure}

3D hand joint estimation is a computer vision challenge that have been addressed through deep learning models in the last years \cite{asadi2017surveyhands}. However, datasets that provide hand pose or joints precise position in 3D are not very common and extensive. The problem consist of estimating hand joint positions in 3D from an RGB image, so data such as segmentation masks, depth (to compute point clouds) and skeleton pose are needed. In this context, UnrealROX+ is pretty useful for generating this kind of data easily from \gls{ue4}. That has been done in \cite{castro2020grasphands}, where a hand pose dataset has been developed using an early version of UnrealROX+ (see Figure \ref{fig:grasp_dataset}). Moreover, a graph convolutional network fed with point cloud data, GraphHands, is presented to validate the generated data.

{}

\subsection{6D pose estimation}

Another widely used technique for which data generated with our tool can be helpful is 6D pose estimation of objects from 2D RGB images. This approach takes the object location problem one step further since it infers 3D rotation of the detected objects besides its location in an image (traditionally represented with a 2D bounding box). As a result, this estimation gives back a 3D bounding box that will estimate both 3D location (centroid) and rotation of the object.

\begin{figure}[htbp]
    \centering
    \includegraphics[width=0.49\linewidth]{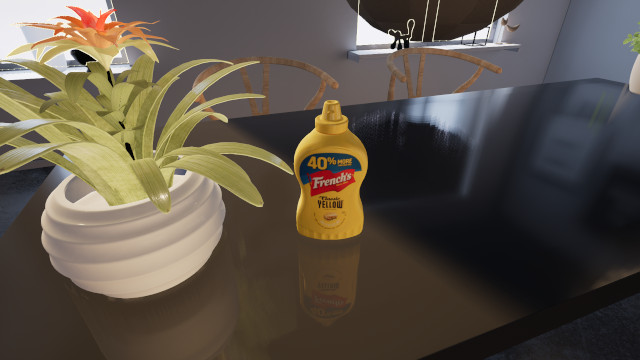}
    \includegraphics[width=0.49\linewidth]{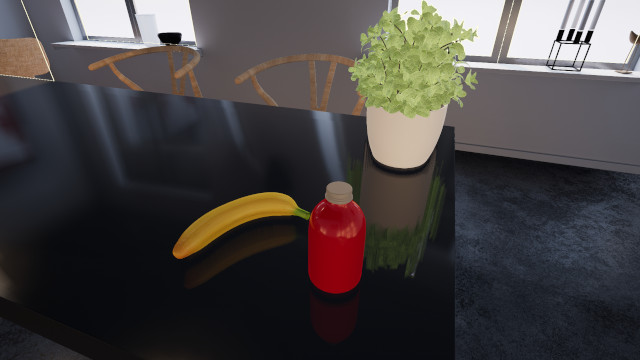}
    \includegraphics[width=0.49\linewidth]{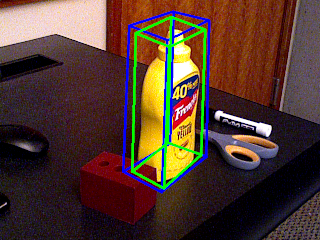}
    \includegraphics[width=0.49\linewidth]{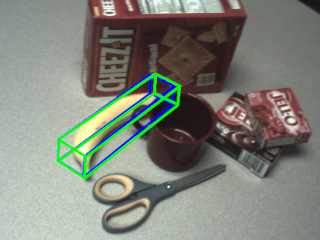}
    \caption{Above, example of training synthetic data with two different objects generated with UnrealROX+. Below, qualitative evaluation of 6D pose estimation with Pix2Pose \cite{Park2019pix2pose} over real data from YCBv dataset \cite{hodan2018bop} (in blue estimated pose, in green ground truth).}
    \label{fig:pix2pose}
\end{figure}

Pix2Pose \cite{Park2019pix2pose} is a recent model that addresses this problem. It implements an auto-encoder architecture designed to estimate 3D position of each pixel from a 2D RGB image. In addition, it includes some interesting particularities, such its robustness to occlusions by leveraging recent achievements in generative adversarial training to precisely recover occluded parts. To prove the usefulness of our generator in this problem, we trained Pix2Pose with our simulated data for single object pose estimation. The objects chosen for the experimentation have been taken from YCB model set \cite{calli2015ycb}, since it provides both real objects, that can be ordered for physically having them, and high-precise registered models from that objects, that can be imported to \gls{ue4}. The real data we are going to use is the one provided from the YCBv dataset, from the BOP 6D pose Benchmark \cite{hodan2018bop}, which contains image sequences with real objects from YCB model set. So, we can generate fully annotated synthetic data to train the network, and real data to test it. Some qualitative results can be appreciated in Figure \ref{fig:pix2pose}, where we can see how 3D bounding boxes for this two objects are pretty well estimated over real data.
%We have tried several configurations for training, including using both real and synthetic data for training. The results, slightly improving some metrics comparing with the ones that have only been trained with real data. Nevertheless, reality gap has to be overcome, and if propor

%% file: sections/conclusion.tex
\section{Conclusion}
\label{sec:conclusion}
This paper presented major improvements on the original UnrealROX, our previously presented tool for generating synthetic data from realistic 3D environments in Unreal Engine 4. It already demonstrated its potential for simulating robotic interactions and generating huge amounts of data that facilitates training data-driven methods for various applications such as semantic segmentation, depth estimation, or object recognition. However, it lacked from the flexibility and modularity for making the tool profitable on a wider range of scenarios, and for a bigger number of researches. So, we worked on decoupling the main data-retrieval system from the main workflow in order to make it much easier and faster to set up and script custom behaviours with Unreal Engine's visual scripting language.
Moreover, new kinds of data, such as albedo or shading maps, or more efficient ways of acquire and managing data, such as segmentation masks or skeleton pose retrieval, were also added, along with useful features such as the Python \gls{api}, useful for Reinforcement Learning.
All UnrealROX original features (multicamera, bounding boxes, ...), and also the original workflow that was used for generating RobotriX, are included or migrated to this new version of the tool, this time as an easy-to-use Unreal Engine plug-in. Grasping system, that has been moved to a separate project that will have its own plug-in, which will be completely compatible with UnrealROX+.

%% file: sections/future.tex
\section{Limitations and Future Works}
\label{sec:future}
Most of the limitations and future works that we presented in our previous paper \cite{Martinez2019unrealrox} have been addressed in this work, and many others are left outside for having decoupled grasping from the data-acquiring tool.
One that remains unsolved is the possibility of managing non-rigid objects and deformations offline, this is, being able to recreate these behaviours in a frame-by-frame basis for acquiring data at a slower and more precise regime. 
%Nevertheless, it is slightly out of scope since it would apply to deformations when interacting or grasping objects, or to pure physics simulations that we rely on \gls{ue4}.
Another possible improvement points are creating a system able to process \gls{URDF} to automatically import robots, 
%, including their constraints, kinematics, and colliders, in the environment instead of doing that manually for each robot model.
or providing an additional segmentation layer for objects behind transparent materials.
Camera distortion and noise are aspects that may distinguish real from synthetic images. They can be addressed as a post-process, but it could be something interesting to parameterize in the tool itself. Other phenomenons, such as camera shake, very common on robotic applications, are not provided or guided by our tool, although they could be simulated through \gls{ue4}.
Talking about new functionalities, the presented Python \gls{api} for Reinforcement Learning is somewhat preliminary and many more commands and more efficient or comfortable ways of working may be added. Lastly, the main future work is definitely the improved and encapsulated version of \textit{UnrealGrasp}, including detailed interaction information.